\documentclass{Interspeech}

\interspeechcameraready

\usepackage[dvipsnames]{xcolor}
\usepackage{multirow}
\usepackage{float,caption}
\usepackage{subcaption}
\usepackage[mode=buildnew]{standalone}

\newcommand{\HermesModel}{\texttt{Hermes-3-Llama-3.1-8B}}

\title{Leveraging Information Retrieval to Enhance Spoken Language Understanding Prompts in Few-Shot Learning}

\author[affiliation={1,2}]{Pierre}{Lepagnol}
\author[affiliation={1}]{Sahar}{Ghannay}
\author[affiliation={1}]{Thomas}{Gerald}
\author[affiliation={1}]{Christophe}{Servan}
\author[affiliation={1}]{Sophie}{Rosset}

\affiliation{}{Université Paris-Saclay, CNRS, LISN}{France}
\affiliation{}{SCIAM}{France}
\email{firstname.lastname@lisn.fr}
\keywords{SLU, Information Retrieval, Prompt Engineering, Few-Shot Learning}

\usepackage{comment}

\begin{document}
\maketitle
\begin{abstract}

Understanding user queries is fundamental in many applications, such as home assistants, booking systems, or recommendations.
Accordingly, it is crucial to develop accurate Spoken Language Understanding (SLU) approaches to ensure the reliability of the considered system.
Current State-of-the-Art SLU techniques rely on large amounts of training data; however, only limited annotated examples are available for specific tasks or languages.

In the meantime, instruction-tuned large language models (LLMs) have shown exceptional performance on unseen tasks in a few-shot setting when provided with adequate prompts.
In this work, we propose to explore example selection by leveraging Information retrieval (IR)  approaches to build an enhanced prompt that is applied to an SLU task.
We evaluate the effectiveness of the proposed method on several SLU benchmarks.
Experimental results show that lexical IR methods significantly enhance performance without increasing prompt length.

\end{abstract}

\begin{figure*}[h!]
    \centering
    \begin{subfigure}[b]{.3\linewidth}
    \includegraphics[width=\linewidth]{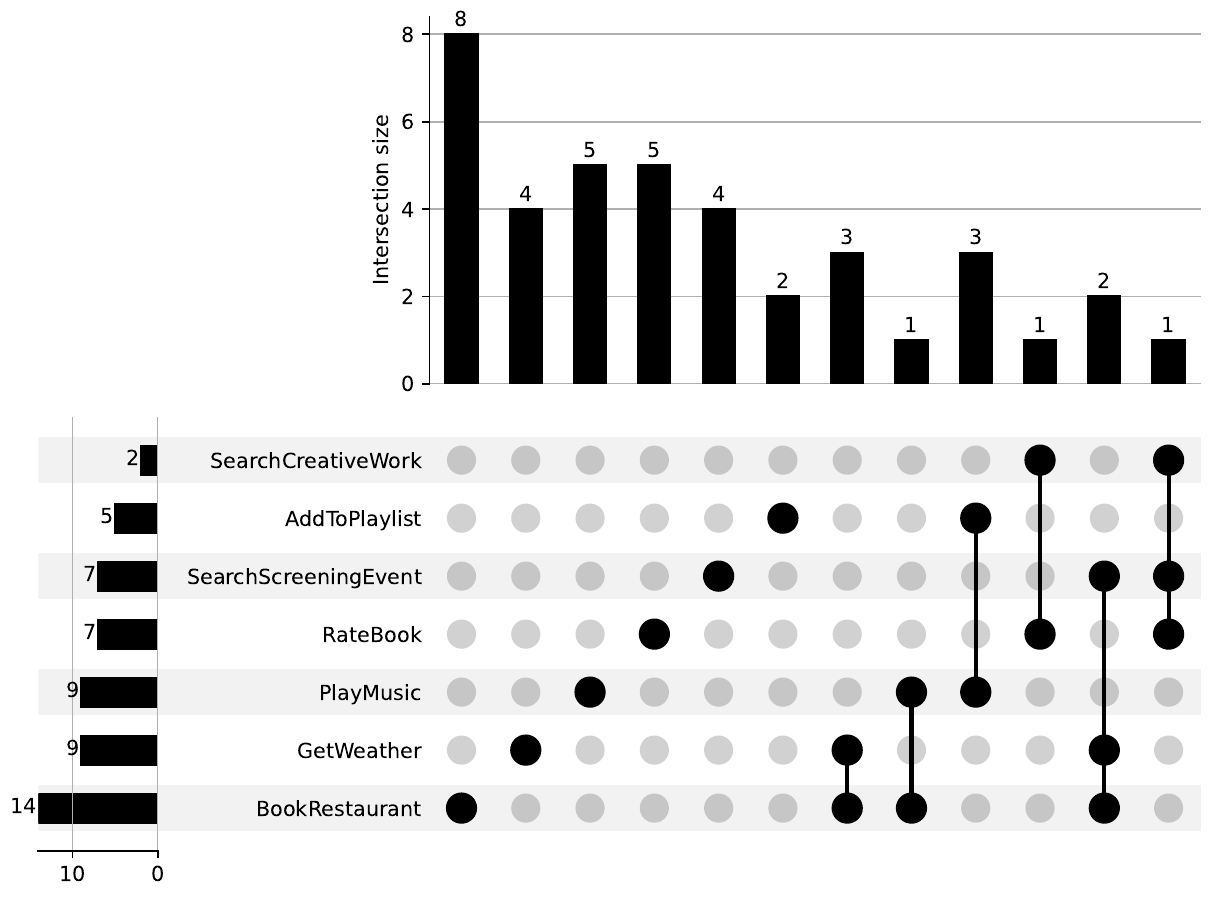}
    \caption{SNIPS}\label{fig:SNIPS_Intents}
    \end{subfigure}
    \hfill
    \begin{subfigure}[b]{.5\linewidth}
    \includegraphics[width=\linewidth]{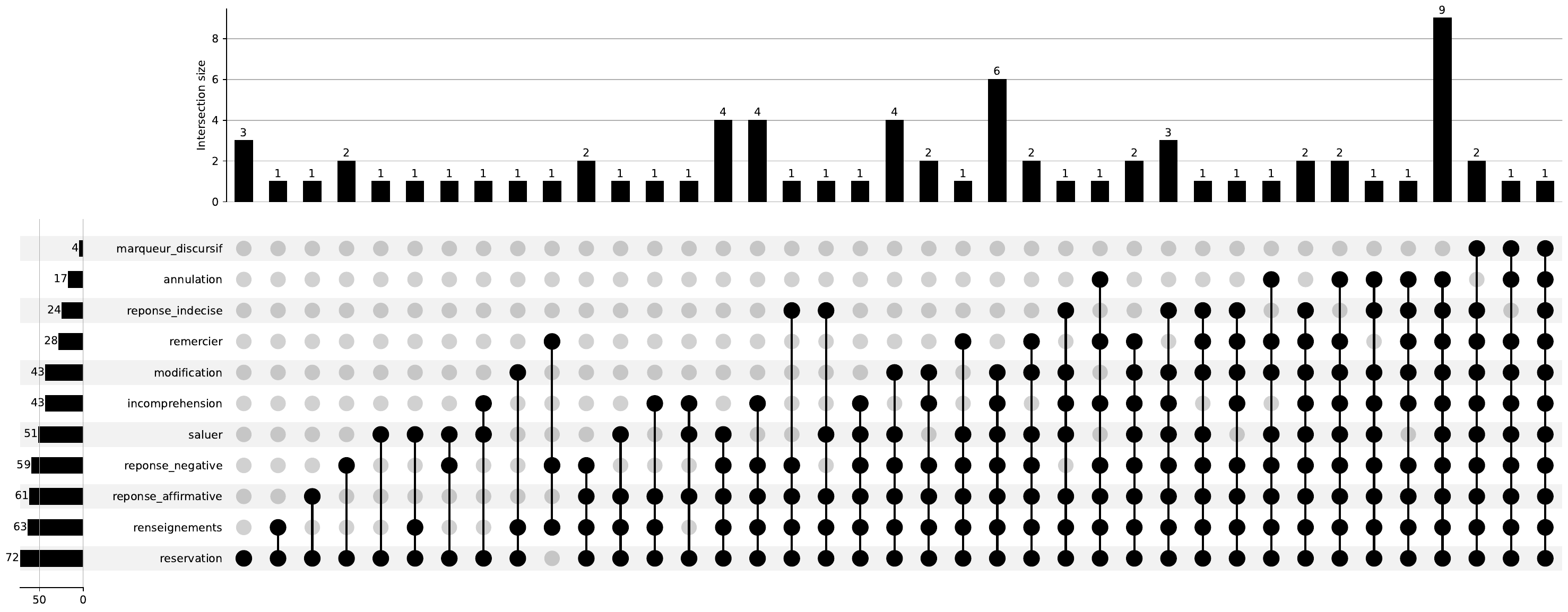}
    \caption{MEDIA}\label{fig:MEDIA_Intents}
    \end{subfigure}

    \caption{Upset plot illustrating the intersections of slots across different intents. In the SNIPS dataset, only a few slots are shared between intents, indicating minimal overlap. In contrast, the MEDIA dataset shows significant overlap, with many slots being shared among multiple intents.}
    \label{fig:IntentSeparation}
    \vspace*{-0.5cm}
\end{figure*}

\section{Introduction}

Spoken Language Understanding (SLU) is a critical component of task-oriented dialogue systems \cite{tur2011spoken}.
It consists of extracting information from user utterances and providing faithful information to a more extensive system.
The extracted information is crucial to the dialogue systems for different sub-tasks, from querying a database (e.g., retrieving items from a database) to answering the user's request.

The SLU task can fall into three sub-tasks: domain classification, i.e. retrieve the domain of an utterance (music recommendation, hotel booking \dots); Intent Classification (IC), which identifies the user's intent (e.g. hotel search, book a hotel, play music, \dots);  Slot-Filling (SF)~\cite{tur2011spoken} where it aims to extract semantics concepts (e.g., date, location, \dots). In this study, we are interested in the Slot-Filling task that can also be considered as a concept detection task~\cite{CI_Bonneau-Maynard2006}. To illustrate the tasks, let's assume the following example:
\textbf{``What is the weather like in Abu Dhabi tomorrow?''}.
From this sentence, the system could extract an intent label (e.g., weather\_information) and the different semantic information, such as \texttt{location:Abu Dhabi} and \texttt{date:tomorrow}.

In recent years, SLU benchmarks and datasets have emerged to assess the relevance of developed approaches. Some are annotated on textual interactions between humans (with an user playing the role of the system) like ATIS~\cite{hemphill-etal-1990-atis} while others are based on automatic transcription, such as MEDIA~\cite{bonneaumaynard05_interspeech} or SLURP\cite{bastianelli-etal-2020-slurp}.

Since the mid-2000s, machine-learning supervised approaches have been widely used to perform SLU tasks, specifically SF tasks \cite{servan:hal-01160181,2010CRF}.
Then, deep-learning approaches have enabled to reach a new step in SLU quality, for instance, mixing LSTM, CNNs and CRF approach~\cite{ma-hovy-2016-end}.
These last years, the fine-tuning of pre-trained transformer-based models (BERT) models~\cite{devlin-etal-2019-bert} was a quality game changer~\cite{ghannay:hal-03007482}.
Today's approaches mainly rely on transformer pre-trained models, tackle either IC and SF independently~\cite{castellucci2019multilingual} or jointly \cite{chen2019bert,zhang2019joint}.
While reaching the best performances, fine-tuned BERT-based approaches suffer from the number of annotated data needed to adapt the model~\cite{weld_survey_2023} to a specific scenario.
Recent approaches have focused on exploiting LLM internal knowledge to perform adaptation with limited number of examples (\textit{few-shot} setting)~\cite{He2023,Mirza2024} or no examples (\textit{zero-shot} setting)~\cite{zhu-etal-2024-zero-shot,qin2024cropromptcrosstaskinteractiveprompting} by prepending the prompts with slots definition.
In the meantime, generative LLM is particularly efficient in mimicking the example scheme given in a prompt. In a few-shot setting, state-of-the-art approaches focus on example selection based on the user's intent~\cite{He2023,qin2024cropromptcrosstaskinteractiveprompting}.
While such approaches are promising, they rely on randomness to select prompt examples.
However, we hypothesize that random prompt examples are not always relevant for a specific utterance~\cite{RAG_CODE_PROMPT}.




 This work aims to study different example selection methods and build an enhanced prompt using only a limited number of relevant examples.
The proposed methodology includes the following:
\begin{itemize}
    \item Exploration of Information Retrieval approaches for SLU tasks to address the challenge of prompt efficiency. This will involve a comparison of different retrieval mechanisms such as BM25~\cite{BM25_and_Beyond} (Best Matching 25), which is a widely-used lexical ranking function and a state-of-the-art contextual embedding approach such as ColBERT model ~\cite{khattab2020colbertefficienteffectivepassage,santhanam2022colbertv2effectiveefficientretrieval}.
    \item Evaluation of our example selection method across several SLU benchmarks (ATIS, SNIPS, SLURP, MEDIA).
\end{itemize}
In order to demonstrate the relevance of the present contributions, the paper is organised as follows: in section~\ref{section:proposed_approach}, the proposed approach is explained, 
in section~\ref{section:experimental_setup} we describe the experimental protocol with the different models, corpora, and configurations; the results of experiments are analyzed in section~\ref{section:result_analysis}; and finally, we conclude in section~\ref{section:conclusion} and discuss limitations of the approach in section~\ref{section:limitation}.

\section{Proposed Approach}
\label{section:proposed_approach}



While selecting the right examples is a key challenge in order to enhance the quality of the prompt \cite{RAG_CODE_PROMPT}, to the best of our knowledge, the current state of the art concerning SLU tasks using prompt engineering for slot-filling (SF) does not select examples based on utterance similarity.
In contrast, related approaches select examples randomly, using only the intention as an anchor~\cite{He2023,qin2024cropromptcrosstaskinteractiveprompting}.
However, an analysis of the SF discrepancy linked to their intent reveals that the slots are not exclusive to intent.
In some cases, the slots overlap significantly across several intents, as illustrated in Figure~\ref{fig:IntentSeparation}.
Restricting examples in prompt to those with identical intent may lead to the exclusion of utterances with the potential to improve performance.
Consequently, it would be advantageous to select examples using other features such as utterance text (or/and semantic) similarity.


To better select examples for the SLU prompting, our approach uses information retrieval (IR) methods.
The proposed pipeline consists of three main steps, detailed as follows: 

\begin{description}
    \item[Querying \& Retrieving Examples] For a given IR method, we use only the utterance to be analyzed to select a set of top-$K$ closest examples.
    This selection is made by computing a similarity score between each utterance example from the train set $U_t$ and the utterance to be analyzed $U_a$.


 \item[Prompt Formatting] The prompt is formatted by including the retrieved examples $U_t$ and their intent to extract slot values using a template (see Figure \ref{fig:PromptTemplate}).

\begin{figure}[h]
    \centering
    \includegraphics[width=0.8\columnwidth]{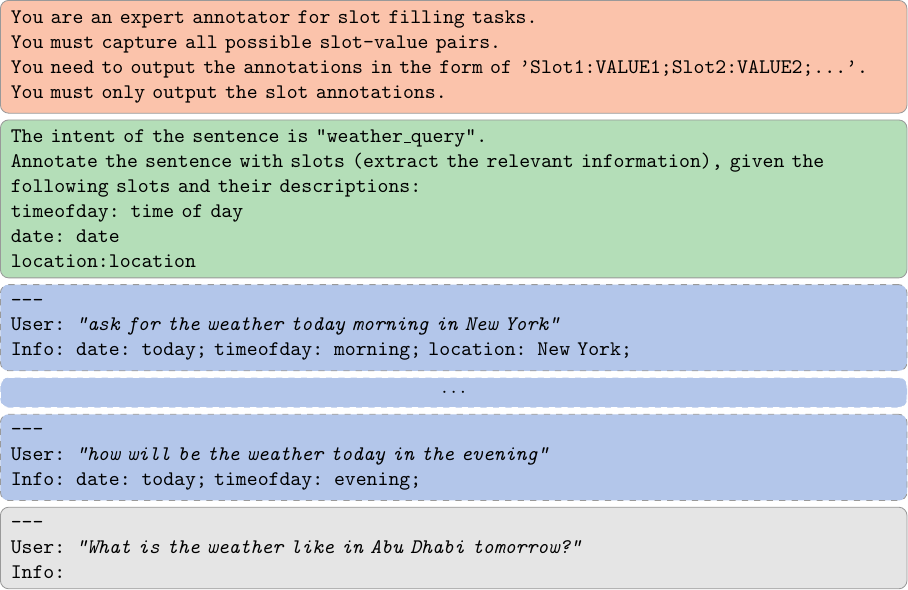}
    \caption{Prompt Template:  In red, the context. In green are the instructions. In blue, examples with labeled slots. In gray is the current utterance we want to label.}
   \label{fig:PromptTemplate}
   \vspace{-0.2cm}

\end{figure}

\item[Generating \& Parsing] The formatted prompt is fed into the LLM after applying the proper prompt template to generate 256 tokens using greedy decoding.
The response is finally parsed using regular expressions that catch each slot type/value pair.



\end{description}

\section{Experimental Setup}
\label{section:experimental_setup}

This section outlines the experimental methodology for evaluating our approach.
We present the datasets used, describe the chosen language model, detail the retrieval mechanisms employed, and specify the baseline and comparison methods.

\subsection{Datasets \& Metrics}

Our study focuses on four well-known SLU benchmarks (see details in table~\ref{tab:DatasetMonolingualStats}), evaluated using the F1-score:
\begin{table}[h!]
\resizebox{\columnwidth}{!}{
    \begin{tabular}{lcccccccc}
    \toprule
    & \multicolumn{2}{c}{ATIS} & \multicolumn{2}{c}{SLURP} & \multicolumn{2}{c}{SNIPS} & \multicolumn{2}{c}{MEDIA} \\
    & Train & Test & Train & Test & Train & Test & Train & Test  \\
    \cmidrule(lr){2-3} \cmidrule(lr){4-5} \cmidrule(lr){6-7} \cmidrule(lr){8-9}
    Nb. utterances & 4978 & 893 & 11514 & 2974 & 13084 & 700 & 13712 & 3767  \\
    Nb. intents & 17 & 16 & 91 & 77 & 7 & 7 & 11 & 11  \\
    Nb. slots & 80 & 70 & 55 & 53 & 39 & 39 & 74 & 72  \\
    \bottomrule
    \end{tabular}
}
\caption{Key characteristics of the datasets: number of utterances, the number of intents, and the number of slots in both the training and test sets.}
\label{tab:DatasetMonolingualStats}
\vspace*{-0.5cm}
\end{table}
\begin{description}
    \item [ATIS:]
 is an English dataset with queries about airline travel information~\cite{hemphill-etal-1990-atis}.
    \item [SNIPS:] is an English dataset from the SNIPS personal assistant~\cite{coucke2018snips}.
    \item [SLURP:] is an English dataset simulating single-turn interactions between users and a voice-controlled assistant~\cite{bastianelli-etal-2020-slurp}.
    \item [MEDIA:] is a French dataset about hotel reservations and information~\cite{bonneaumaynard05_interspeech, alavoine:hal-04523286}. MEDIA corpus is available in a \textit{full} or a \textit{relax scoring} version.  In this study, we used the 2022 version~\cite{Laperriere2022} of MEDIA with the relaxed scoring version, in which attributes are simplified by excluding the specifiers. This version was enriched with intent labels by~\cite{alavoine:hal-04523286}.

\end{description}

\subsection{Language Model}

We selected an instruction/chat fine-tuned version of Llama 3.1 8B~\cite{dubey2024llama3herdmodels} model called \HermesModel{} \cite{teknium2024hermes3technicalreport} for the following reasons: 1)  Its long context window allowing to consider long sequences of tokens as input (numerous examples); 2) Its ability in generating structured outputs, making it an appropriate choice for our highly structured task.




\subsection{Example selection approaches}
\label{subsec:selection}

To evaluate our example selection approach within the whole training dataset, we propose to compare the two following retrieval methods: the classical lexical \textbf{BM25} \cite{BM25_and_Beyond} and the state-of-the-art contextual embedding approach \textbf{ColBERT} \cite{khattab2020colbertefficienteffectivepassage,santhanam2022colbertv2effectiveefficientretrieval} (see Appendix in supplementary materials for more details).
We add the \textbf{Random} setup, which consists of taking a random selection of examples from the whole example set, and the
\textbf{Intent-based} selection takes random examples from the examples set within the same intent.
This last method is largely used in previous work~\cite{zhu-etal-2024-zero-shot,Mirza2024};
We also filter our selected examples based on the intent, before retrieval, named \textbf{Intent$\rightarrow$BM25} and the reverse: BM25 before intent selection
(noted \textbf{BM25$\rightarrow$Intent}).

Following \cite{Mirza2024}, we compare our method (prompt template + example selection method) within the few-shot in-context learning setting using 10 examples.
As their experience is partially reproducible (only a manual sub-sample of the training set was considered), we use their prompt template and their selection examples to ensure a fair comparison denoted as \texttt{Hermes-3 [Recomputed]}\footnote{Note that their best model is no longer available, which led us to use the \HermesModel~ model instead}.


We also reported F1-scores from fully fine-tuned models. Namely HAN (Higher-order Attention Network)~\cite{ijcai2022p565}, FlauBERT-oral-ft~\cite{alavoine:hal-04523286} and HERMIT\cite{vanzo-etal-2019-hierarchical} a Bi-LSTM with CRF model. They represent the current SOTA for the selected SLU Benchmarks.

\section{Results \& analysis}
\label{section:result_analysis}
In this section, we report and discuss the results of the different experiments.
It includes a comparison to state-of-the-art SLU methods, the evaluation of the different example selection methods, the impact of retrieved utterances on performances, and the impact of the number of selected examples.

\subsection{Comparative analysis with State-of-the-Art Methods}


Table~\ref{tab:FewShotPerformancesSOTA} presents a comparison between the proposed approaches and state-of-the-art.
The results, measured in F1-scores, showcase three main categories of models: fully fine-tuned, few-shot prompting with 10 examples from ~\cite{Mirza2024} and the proposed approaches using \texttt{Hermes-3-Llama-3.1-8B} model.

Fully fine-tuned models, particularly HAN, FlauBERT-oral-ft, and HERMIT, are the topline and demonstrate superior performance on all datasets, achieving the highest F1-scores on ATIS (97.23), MEDIA (87.75),  SNIPS (98.26) and SLURP (78.19) \cite{alavoine:hal-04523286,ijcai2022p565,vanzo-etal-2019-hierarchical, bastianelli-etal-2020-slurp}.
Our baseline uses the experiments from \cite{Mirza2024} using the \texttt{Hermes-3-Llama-3.1-8B} model, which reaches 83.98 of F1-scores.
This approach uses intent selection from a manually selected pool of examples from the SNIPS corpus.
Since it is not possible to reproduce this manual selection for every corpus, we propose to select the utterances following the approaches we presented in section \ref{subsec:selection}.
The results of \texttt{flan-t5-xxl} and \texttt{WizardLM-13B-V1.1} come from \cite{Mirza2024}.

Random selection, serving as baseline, achieves only 38.43\%F1, and does not ensure that the retrieved examples align with the specific query.
In contrast, intent-based selection, which is the closest to \cite{Mirza2024} approach, improves performance to 55.23 points in F1 by ensuring that examples share the same high-level intent.
Yet, it overlooks fine-grained lexical or slot-level similarities, crucial for this task.

According to the F1 scores, BM25-based methods consistently outperform other selection methods. Notably, the BM25 method is enhanced by incorporating intent information.
The ``BM25 $\rightarrow$ Intent'' approach performed similarly to the ``Intent $\rightarrow$ BM25'' method. However, on ATIS and MEDIA it averaged about 0.794 points lower, while on SLURP and SNIPS the gap increased to roughly 1.5 points.

Finally, the ColBERT approach lies behind BM25 with 55.73 points and 67.61 points, respectively.
It indicates that semantic information is less effective than lexical information in selecting examples for prompting.
We can conclude that the lexical alignment between utterances and slots is the most valuable information to pass on to the LLM.

\begin{table}[h]
  \centering
    \resizebox{\columnwidth}{!}{
\begin{tabular}{lccccc}
    \toprule
  Model \& Selection  & ATIS & MEDIA & SLURP & SNIPS & Mean\\
  \midrule
  \midrule
   \multicolumn{6}{c}{Fully Fine-tunned models~(need the full training set)}\\
    FlauBERT-oral-ft~\cite{alavoine:hal-04523286} & - & \textbf{87.75} & - &  - & N/A\\
    HAN~\cite{ijcai2022p565} & \textbf{97.23} & - & \underline{55.50} & \textbf{98.26} & N/A\\

    HERMIT\cite{vanzo-etal-2019-hierarchical, bastianelli-etal-2020-slurp} &-  &- & \textbf{78.19} &- & N/A\\
  \midrule
  \midrule
   \multicolumn{6}{c}{Few-shot Prompting~\cite{Mirza2024} using 10 examples } \\
  flan-t5-xxl & - & - & 47.30 &  64.70& N/A \\
  WizardLM-13B-V1.1  & - & - & 47.40 & 68.50& N/A\\
\texttt{Hermes-3 [Recomputed]} & -  & - & - & 83.98 & N/A\\
  \midrule
  \midrule
   \multicolumn{6}{c}{This work: Few-shot Prompting   using 10 examples}\\
  \texttt{Hermes-3} + Random &  68.24  &  26.55  &  15.20  &  43.73  & 38.43\\
  \texttt{Hermes-3} + Intent-Based &  72.37  &  34.78  &  35.72  &  78.03  & 55.23\\
  \texttt{Hermes-3} + ColBERT  &  83.57  &  45.23  &  24.02  &  70.11 & 55.73 \\
  \texttt{Hermes-3} + BM25  &  86.63  &  55.56  &  43.87  &  84.38  & 67.61\\
  \texttt{Hermes-3} + BM25$\rightarrow$Intent  &  85.11  &  60.78  &  46.14  &  85.21 & 69.31 \\
  \texttt{Hermes-3} + Intent$\rightarrow$BM25  &  86.38  &  61.10  &  44.01  &  84.32  & 68.95\\
\bottomrule
    \end{tabular}
    }
  \caption{Results (F1-scores) in Slot Filling task. In \textbf{bold} the best score, \underline{underline} the second best score.}
  \label{tab:FewShotPerformancesSOTA}
\end{table}
\vspace{-0.5cm}
\subsection{Impact of Retrieved utterances on Performances}
In this section, we examine the impact of different selection methods to identify the usefulness of the retrieved examples.
We hypothesized that the retrieved examples add additional information for the model to use to generate the correct slots.

\begin{table}[H]
\centering
\resizebox{\columnwidth}{!}{
\begin{tabular}{llrr}
\toprule
dataset & method  & Slot Presence & Intents Presence \\
\midrule
\multirow[c]{3}{*}{ATIS} & BM25  & \textbf{89.75\% $\pm$ 7.47\%} & \underline{94.27\% $\pm$ 5.32\%} \\
 & ColBERT & \underline{80.91\% $\pm$ 15.06\%} & 84.95\% $\pm$ 12.41\% \\
 & Intent-Based Selection & 75.59\% $\pm$ 14.06\% & \textbf{99.34\% $\pm$ 0.34\%} \\
 & Random Selection & 67.14\% $\pm$ 16.49\% & 76.70\% $\pm$ 12.46\% \\
\midrule
\multirow[c]{3}{*}{MEDIA} & BM25  & \textbf{90.53\% $\pm$ 7.76\%} & \underline{93.42\% $\pm$ 5.38\%} \\
 & ColBERT & 65.75\% $\pm$ 22.74\% & 72.49\% $\pm$ 22.37\% \\

 & Intent-Based Selection & \underline{75.61\% $\pm$ 16.87\%} & \textbf{98.18\% $\pm$ 2.02\%} \\
 & Random Selection & 65.00\% $\pm$ 23.23\% & 68.05\% $\pm$ 26.71\% \\
\midrule
\multirow[c]{3}{*}{SLURP} & BM25  & \textbf{89.36\% $\pm$ 8.83\%} & \underline{87.49\% $\pm$ 9.95\%} \\
 & ColBERT & 59.53\% $\pm$ 15.15\% & 44.21\% $\pm$25.73\% \\
 & Intent-Based Selection & \underline{83.37\% $\pm$ 14.15\%} & \textbf{99.97\% $\pm$ 0.00\%} \\
 & Random Selection & 57.76\% $\pm$ 21.41\% & 28.87\% $\pm$ 32.80\% \\
\midrule
\multirow[c]{3}{*}{SNIPS} & BM25  & \textbf{89.81\% $\pm$ 11.49\%} & \underline{96.41\% $\pm$ 5.14\%} \\
 & ColBERT & 72.77\% $\pm$ 21.28\% & 88.73\% $\pm$ 10.92\% \\
 & Intent-Based Selection & \underline{83.52\% $\pm$ 17.86\%} & \textbf{100.00\% $\pm$ 0.00\%} \\
 & Random Selection & 52.18\% $\pm$ 35.10\% & 58.08\% $\pm$ 35.67\% \\
\bottomrule
\end{tabular}
    }
\caption{Comparison of the presence of expected slots and intents between each selection method. For example, in ATIS for BM25, 89.75\% of the prompt have the right and expected slots into the prompt. In \textbf{bold} the best score, \underline{underline} the second best score.}
\label{tab:PresenceConceptsIntents}
\end{table}
\vspace{-0.5cm}
\begin{figure*}[ht]
        \centerline{\includegraphics[width=0.9\textwidth]{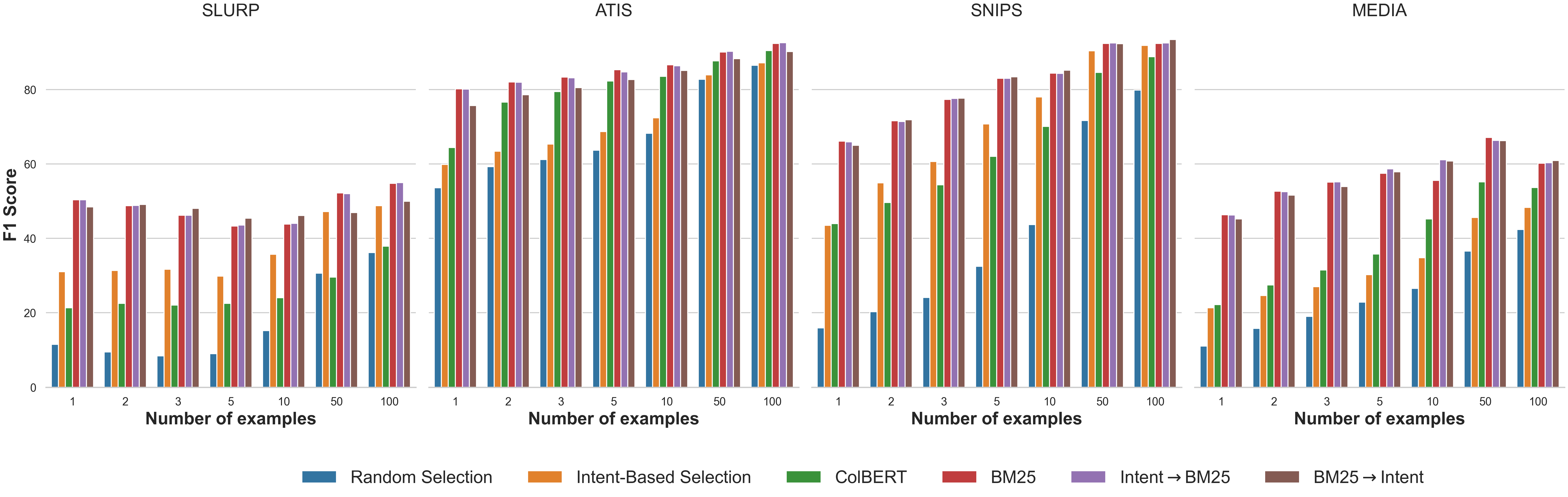}}
    \caption{Performance of the different methods for each number of examples in the prompt for ATIS, SLURP, SNIPS, and MEDIA}
\label{fig:FewshotsPerformances_per_examples}
\end{figure*}

Table~\ref{tab:PresenceConceptsIntents} presents the percentage of examples where each method retrieves correct intents and slots (on average over the number of examples).
We observe that, BM25 selects more frequently examples containing the necessary slots compared to Intent-based Selection or Random selection, hence it achieved the best score F1-score.
BM25 misses the correct slots in only 10\% of the test set against about 20-25\% for Intent-based methods.
On the MEDIA dataset, 90.53\% of the prompts include the appropriate slots versus 75.61\% for Intent-Based Selection.
Furthermore, BM25 selects examples based on the words in the utterance, meaning the selected examples are very close to the current utterance we want to analyze.

The semantic retrieval method with ColBERT underperforms BM25 method and often Intent-Based Selection (on SNIPS and SLURP).
In the same way, ColBERT miss more often the appropriate slots compared to
Intent-Based Selection. 

In summary, the results demonstrate that retrieval methods integrating BM25, especially when combined with intent information (either before or after retrieval), substantially outperform random retrieval approaches.
The optimal configuration appears to be dataset-dependent: BM25 excels in datasets like ATIS, where slot types are less varied, while intent-guided BM25 methods perform better in more complex datasets like MEDIA and SLURP.
This could suggest that tailoring the retrieval strategy to the specific characteristics of the dataset can lead to significant improvements in Slot Filling performance or some redundancies in the ATIS dataset make it easier.
Nevertheless, as ATIS was published in the 1990s, it is possible that it was subsequently leaked onto the internet and incorporated into the training set of the \HermesModel{} or Llama 3.1 models.

\subsection{Varying the number of examples}
We study the impact of the number of examples in prompts across the selection approaches and the benchmarks used.
This aims to determine what could the right amount of data needed to perform SLU with the \HermesModel{}.

Figure~\ref{fig:FewshotsPerformances_per_examples} presents the performance of each selection method across the different datasets by varying the number of examples in the prompt.
Using BM25 methods (BM25, BM25 $\rightarrow$ intent and, intent $\rightarrow$ BM25) consistently outperforms the others methods. Particularly, considering only five examples using BM25 always outperforms random methods considering 100 examples. Consequently, retrieval-based prompts improve performances while benefiting from a lower computational cost (limiting the input sequence length and thus lowering the cost of self-attention processing).
While Intent-Based selection significantly improves at higher shot counts, it never surpasses BM25-based methods. These findings suggest that leveraging BM25 as a foundation for slot-filling provides a significant advantage in scenarios with limited labelled examples.


\section{Conclusion}
\label{section:conclusion}

This study investigates the use of information retrieval (IR) methods to enhance prompt construction in spoken language understanding (SLU) task.

By integrating IR methods, specifically BM25, into the example selection process, we have addressed the challenges of overlapping intents and slots in complex datasets.  Our comprehensive evaluation across multiple benchmarks showed that BM25 consistently improved F1 performance scores over traditional intent-based selection methods.
Moreover, these improvements were achieved without increasing prompt length.
We investigated the impact of the number of selected examples and compared both lexical and semantic IR methods.
BM25-based methods showed consistent efficiency in different SLU Benchmarks. 
The effectiveness of the approach is highly dependent on the quality of the retrieval mechanism;
suboptimal retrieval could adversely affect overall performance, as demonstrated by the results of ColBERT, which failed to retrieve relevant examples.
While the method maintains similar prompt lengths, the retrieval process can introduce additional computational overhead, with larger datasets, which can be significant in real-time applications where latency is critical.

In conclusion, BM25 represents a step forward in the field of prompting for SLU tasks.
By effectively balancing specificity and length, and improving performance without additional computational cost, our method offers a practical solution to improve the accuracy and applicability of SLU systems across diverse linguistic and industrial contexts.

To ensure reproducibility, the code and resources used in this study are available at \url{https://gitlab.lisn.upsaclay.fr/phd-pierre-lepagnol/ir-for-slu}.

\section{Limitations}
\label{section:limitation}

Despite promising advances, this study on language models for SLU tasks may have several limitations.
Firstly, the reliance on access to a training dataset, for example retrieval, poses challenges, particularly in low-resource settings where such data may be scarce or unavailable, thus limiting the applicability of the method.
In addition, there is a possibility that some models performed well because they were trained on all or a part of our considered datasets, namely ATIS and SNIPS.
The study also used a single instruction prompt without exploring variations of the task formulation, potentially limiting the understanding of how different prompt designs affect performance.
Furthermore, the study focused on one model, potentially overlooking other competitive models.
Finally, the robustness of the method in dealing with rare or highly ambiguous slots has not been thoroughly explored, and inaccuracies in initial intent classification could lead to the selection of inappropriate examples, thereby degrading the performance of the system.
Future work will focus on addressing these limitations.

\section{Acknowledgments}
This work is supported by the ANRT (Association nationale de la recherche et de la technologie) with
a CIFRE fellowship granted to SCIAM\footnote{\url{https://www.sciam.fr/}} (CIFRE N°2022/1608).

This work was performed using HPC resources from GENCI-IDRIS (Grant 2023-AD011014242).

\bibliographystyle{IEEEtran}
\bibliography{fullbib}
\end{document}